\documentclass[letterpaper, 10 pt, conference]{ieeeconf}  
\IEEEoverridecommandlockouts                                                                 
\usepackage{amsmath,amssymb,amsfonts}
\usepackage{graphicx}
\usepackage{textcomp}
\usepackage{xcolor}
\usepackage{lipsum}
\usepackage{svg}
\usepackage{comment}
\def\BibTeX{{\rm B\kern-.05em{\sc i\kern-.025em b}\kern-.08em
    T\kern-.1667em\lower.7ex\hbox{E}\kern-.125emX}}

\usepackage{comment}

\usepackage{nohyperref}

\definecolor{matplotlib0}{HTML}{1f77b4}
\definecolor{matplotlib1}{HTML}{d62728}
\definecolor{matplotlib2}{HTML}{2ca02c}
\definecolor{matplotlib3}{HTML}{ff7f0e}
\definecolor{matplotlib4}{HTML}{9467bd}
\definecolor{matplotlib5}{HTML}{8c564b}
\definecolor{matplotlib6}{HTML}{e377c2}
\definecolor{matplotlib7}{HTML}{7f7f7f}
\definecolor{matplotlib8}{HTML}{bcbd22}
\definecolor{matplotlib9}{HTML}{17becf}

\usepackage{mathtools}

\usepackage{subcaption}

\usepackage{booktabs}
\usepackage{array}
\usepackage{multirow}
\usepackage{colortbl}
\usepackage{tablefootnote}
\usepackage{threeparttable}

\usepackage[acronym, style=super, nonumberlist]{glossaries}

\usepackage{pgfplots}
\definecolor{color0}{rgb}{0.12156862745098,0.466666666666667,0.705882352941177} 
\definecolor{color1}{rgb}{1,0.498039215686275,0.0549019607843137}
\definecolor{color2}{rgb}{0.172549019607843,0.627450980392157,0.172549019607843} 
\definecolor{color3}{rgb}{0.83921568627451,0.152941176470588,0.156862745098039} 
\definecolor{color4}{rgb}{0.580392156862745,0.403921568627451,0.741176470588235}
\definecolor{colorblue}{rgb}{0.12156862745098,0.466666666666667,0.705882352941177} 
\definecolor{colorgreen}{rgb}{0.172549019607843,0.627450980392157,0.172549019607843} 
\definecolor{colorred}{rgb}{0.83921568627451,0.152941176470588,0.156862745098039} 
\definecolor{colorblack}{rgb}{0,0,0} 
\definecolor{colororange}{rgb}{1,0.56,0} 
\usepgfplotslibrary{fillbetween}
\usepgfplotslibrary{colormaps}
\pgfplotsset{compat=1.16}

\pgfplotscreateplotcyclelist{matplotlib}{
  {matplotlib0},
  {matplotlib1},
  {matplotlib2},
  {matplotlib3},
  {matplotlib4},
  {matplotlib5},
  {matplotlib6},
  {matplotlib7},
  {matplotlib8},
  {matplotlib9}
}

\pgfplotsset{every axis/.append style={
    cycle list name=matplotlib
}}

\usepackage{listings}

\definecolor{code_default}{HTML}{000000}
\definecolor{code_keyword}{HTML}{AC4142}
\definecolor{code_identifier}{HTML}{D28445}

\usepackage{algorithm}
\usepackage{algpseudocode}
\usepackage{float}
\newfloat{algorithm}{t}{top}

\newacronym{simd}{SIMD}{Single Instruction, Multiple Data}
\newacronym{elu}{ELU}{Exponential Linear Unit}
\newacronym{relu}{ReLU}{Rectified Linear Unit}
\newacronym{rpr}{RPR}{Random Partition Relaxation}
\newacronym{mac}{MAC}{Multiply Accumulate}
\newacronym{dma}{DMA}{Direct Memory Access}
\newacronym{bmi}{BMI}{Brain--Machine Interface}
\newacronym{bci}{BCI}{Brain--Computer Interface}
\newacronym{smr}{SMR}{Sensory Motor Rythms}
\newacronym{eeg}{EEG}{Electroencephalography}
\newacronym{svm}{SVM}{Support Vector Machine}
\newacronym{svd}{SVD}{Singular Value Decomposition}
\newacronym{evd}{EVD}{Eigendecomposition}
\newacronym{iir}{IIR}{Infinite Impulse Response}
\newacronym{fir}{FIR}{Finite Impulse Response}
\newacronym{fc}{FC}{Fabric Controller}
\newacronym{nn}{NN}{Neural Network}
\newacronym{mrc}{MRC}{Multiscale Riemannian Classifier}
\newacronym{flop}{FLOP}{Floating Point Operation}
\newacronym{sos}{SOS}{Second-Order Section}
\newacronym{ipc}{IPC}{Instructions per Cycle}
\newacronym{tcdm}{TCDM}{Tightly Coupled Data Memory}
\newacronym{fpu}{FPU}{Floating Point Unit}
\newacronym{fma}{FMA}{Fused Multiply Add}
\newacronym{alu}{ALU}{Arithmetic Logic Unit}
\newacronym{dsp}{DSP}{Digital Signal Processing}
\newacronym{gpu}{GPU}{Graphics Processing Unit}
\newacronym{soc}{SoC}{System-on-Chip}
\newacronym{mi}{MI}{Motor-Imagery}
\newacronym{csp}{CSP}{Commmon Spatial Patterns}
\newacronym{fbcsp}{FBCSP}{Filter-Bank \acrlong{csp}}
\newacronym{pulp}{PULP}{parallel ultra-low power}
\newacronym{soa}{SoA}{state-of-the-art}
\newacronym{bn}{BN}{Batch Normalization}
\newacronym{isa}{ISA}{Instruction Set Architecture}
\newacronym{ecg}{ECG}{Electrocardiogram}
\newacronym{mcu}{MCU}{microcontroller}
\newacronym{rnn}{RNN}{recurrent neural network}
\newacronym{cnn}{CNN}{convolutional neural network}
\newacronym{tcn}{TCN}{temporal convolutional network}
\newacronym{emu}{EMU}{epilepsy monitoring unit}

\newacronym{ssl}{SSL}{self-supervised learning}
\newacronym{nlp}{NLP}{natural language processing}
\newacronym{gpt}{GPT}{Generative Pre-trained Transformer}
\newacronym{ssm}{SSM}{State Space Model}
\newacronym{moe}{MoE}{Mixture of Experts}

\newacronym{lda}{LDA}{Linear Discriminant Analysis}
\newacronym{sota}{SOTA}{state-of-the-art}
\newacronym{auroc}{AUROC}{Area Under the Receiver Operating Characteristic}
\newacronym{aupr}{AUPR}{Area Under the Precision-Recall}

\makeatletter
\def\ps@IEEEtitlepagestyle{%
  \def\@oddfoot{\mycopyrightnotice}%
  \def\@evenfoot{}%
  \def\@oddhead{}
  \def\@evenhead{}%
}
\makeatother
\def\mycopyrightnotice{%
  \begin{minipage}{\textwidth}
  \centering \scriptsize
  \copyright 2025 IEEE.  Personal use of this material is permitted.  Permission from IEEE must be obtained for all other uses, in any current or future media, including reprinting/republishing this material for advertising or promotional purposes, creating new collective works, for resale or redistribution to servers or lists, or reuse of any copyrighted component of this work in other works.
  \end{minipage}
}
\makeatother
\begin{document}
\title{\LARGE \bf
FEMBA: Efficient and Scalable EEG Analysis with a \\ Bidirectional Mamba Foundation Model
}

\author{
Anna Tegon$^{1}$, Thorir Mar Ingolfsson$^{1}$, \\ Xiaying Wang$^{1}$, 
Luca Benini$^{1,2}$, Yawei Li$^{1}$
\thanks{$^{1}$Integrated Systems Laboratory, ETH Z{\"u}rich, Z{\"u}rich, Switzerland.}
\thanks{$^{2}$DEI, University of Bologna, Bologna, Italy.}
\thanks{Anna Tegon and Thorir Mar Ingolfsson are co-first authors.}
}
\maketitle
\thispagestyle{IEEEtitlepagestyle}
\begin{abstract}
Accurate and efficient electroencephalography (EEG) analysis is essential for detecting seizures and artifacts in long-term monitoring, with applications spanning hospital diagnostics to wearable health devices. Robust EEG analytics have the potential to greatly improve patient care. However, traditional deep learning models, especially Transformer-based architectures, are hindered by their quadratic time and memory complexity, making them less suitable for resource-constrained environments. To address these challenges, we present FEMBA (Foundational EEG Mamba + Bidirectional Architecture), a novel self-supervised framework that establishes new efficiency benchmarks for EEG analysis through bidirectional state-space modeling. Unlike Transformer-based models, which incur quadratic time and memory complexity, FEMBA scales linearly with sequence length, enabling more scalable and efficient processing of extended EEG recordings. Trained on over 21,000 hours of unlabeled EEG and fine-tuned on three downstream tasks, FEMBA achieves competitive performance in comparison with transformer models, with significantly lower computational cost. Specifically, it reaches 81.82\% balanced accuracy (0.8921 AUROC) on TUAB and 0.949 AUROC on TUAR, while a \emph{tiny} 7.8M-parameter variant demonstrates viability for resource-constrained devices. These results pave the way for scalable, general-purpose EEG analytics in both clinical and highlight FEMBA as a promising candidate for wearable applications.
\newline
\noindent\textit{Clinical relevance}—
By reducing model size and computational overhead, FEMBA enables continuous on-device EEG monitoring for tasks like seizure detection and artifact reduction, promising improved patient care through timely and cost-effective neuro-monitoring solutions.
\end{abstract}
    
\section{Introduction}\label{sec:intro}
The emergence of foundation models has profoundly impacted artificial intelligence, bringing forward a shift toward generalizable, large-scale pre-training. These models, trained via \gls{ssl} on heterogeneous datasets, derive their effectiveness from hierarchical feature extraction that can span diverse tasks~\cite{bommasani2021opportunities}. While their success in language (e.g., BERT~\cite{devlin2019bertpretrainingdeepbidirectional}) and vision (e.g., CLIP~\cite{radford2021learning}) is well-documented, their potential in biomedical signal processing—particularly for \gls{eeg}—remains relatively underexplored.

\gls{eeg} is a challenging modality due to its pseudo-random, non-stationary waveforms, susceptibility to artifacts~\cite{ingolfsson_minimizing_2024}, and substantial intra- and inter-subject variability. These factors demand models that balance robustness with interpretability. Wearable EEG devices play a crucial role in enabling continuous brain monitoring in real-world settings, offering new opportunities for brain-computer interfaces~\cite{zhang2023recent}, healthcare and cognitive research~\cite{emish2024remote}. Although recent efforts have used convolutional architectures~\cite{roy2019deep} and attention-based mechanisms~\cite{chen2024eegformer} for \gls{eeg}, real-world constraints complicate their deployment. Wearable devices and continuous monitoring systems impose strict limits on memory and latency~\cite{casson2010wearable}, making even moderately sized Transformers impractical. As a result, there is a strong, important, unmet need for architectures that can combine expressive power with computational efficiency. Given these constraints, we propose harnessing \glspl{ssm}. Specifically, we build upon the Mamba linear \gls{ssm}, a scalable approach tailored for large-scale \gls{eeg}, to mitigate the memory and latency bottlenecks associated with Transformer models while maintaining high performance and interpretability.

\begin{figure*}[t]
    \centering
    \includegraphics[width=0.9\textwidth]{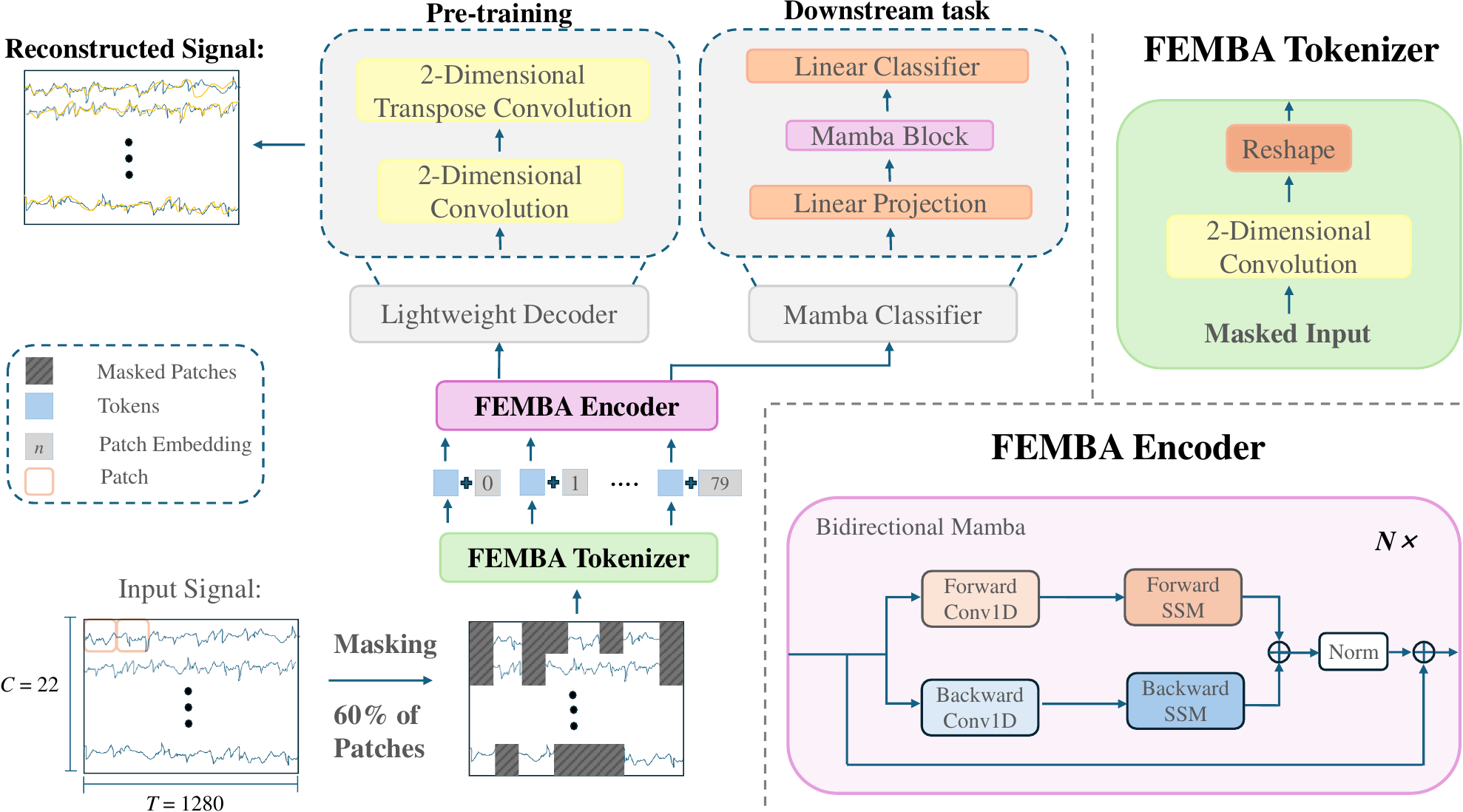}
    \caption{Overview of the proposed FEMBA (Foundational \gls{eeg} Mamba + Bidirectional Architecture) pipeline. The input \gls{eeg} signal (with channels $C$ and length $T$) is first tokenized via a 2D convolution and flattening layer. Random masking is then applied to a subset of the patches for self-supervised learning. The masked tokens pass through the FEMBA encoder, which stacks multiple Bidirectional Mamba blocks. Within each block, the sequence is processed by parallel forward and backward Mamba components (the backward component operating on a temporally reversed input sequence). The outputs from both directions are then combined (e.g., via summation) to capture dependencies from both past and future contexts before potentially passing through normalization and feed-forward layers. Finally, a lightweight decoder (for reconstruction) or a classification head (for downstream tasks) reconstructs or classifies the signals, respectively.}
    \label{fig:femba_architecture}
    \vspace{-0.5cm}
\end{figure*}

\paragraph*{From Transformers to Mamba}
Transformer-based models have demonstrated strong performance in capturing long-range dependencies in \gls{eeg}~\cite{chen2024eegformer,jianglarge}. Current \gls{eeg} foundation models (e.g., BENDR, EEGFormer, LaBraM, Neuro-GPT) predominantly rely on attention mechanisms and may not provide the efficiency demanded by edge-computing environments. However, their $\mathcal{O}(N^2)$ complexity in computation and memory as a function of sequence length N can become a bottleneck for continuous or extended \gls{eeg} recordings, especially on resource-constrained devices. In contrast, Mamba~\cite{gu2023mamba}, which is based on a state-space framework, helps address these challenges by reformulating sequence modeling as a latent differential system. This approach offers linear scaling (as a function of sequence length) without substantially compromising temporal resolution. Bidirectional extensions~\cite{liang_bi-mamba_2024} further enable retrospective analysis, which may be essential for detecting ephemeral biomarkers (e.g., interictal spikes).

To investigate computationally efficient architectures such alternatives, we introduce \emph{\textbf{FEMBA}} (Foundational \gls{eeg} Mamba + Bidirectional Architecture), which leverages state-space principles for large-scale \gls{eeg} modeling. FEMBA is designed to address three key limitations of prior work: (1) quadratic scaling in attention-based models, (2) limited pre-training scope for capturing neurophysiological diversity, and (3) difficulties in adapting to low-resource settings. By pre-training on 21,000 hours of unlabeled \gls{eeg} from 5,000 subjects, FEMBA aims to learn representations that generalize across a range of pathologies, while retaining the potential for deployment on wearable hardware, as demonstrated by the promising performance of our Tiny FEMBA model. We open-source our code for reproducibility\footnote{https://github.com/pulp-bio/BioFoundation}

Our contributions are the following:

\begin{itemize}
    \item \textbf{A Novel Architectural Paradigm:} We integrate a bidirectional state-space approach with \gls{ssl} to demonstrate that linear-time architectures can match—or in some cases surpass—Transformer-based models on established \gls{eeg} benchmarks (TUAB, TUAR, TUSL). This result suggests that attention-based solutions may not always be indispensable for effective \gls{eeg} modeling.
    
    \item \textbf{Large-Scale Pre-training on \gls{eeg}:} We conduct pre-training on a terabyte-scale unlabeled \gls{eeg} dataset (over 21,000 hours of data from more than 5,000 participants) that spans multiple studies. Using random masking for self-supervised reconstruction, FEMBA acquires robust, general representations suitable for diverse downstream tasks without extensive labeled data.
    
    \item \textbf {Efficient \gls{sota} performance} We have developed FEMBA in four sizes of model parameters: Tiny (7.8M), Base (47.7M), Large (77.8M), and Huge (389M). The Huge model achieves a mere 0.7\% decrease in accuracy compared to the \gls{sota} on TUAB while being \textbf{$3.5\times$} more computationally efficient and \textbf{$1.5\times$} more memory efficient. On TUAR, FEMBA sets a new \gls{sota} benchmark, with the Tiny model beating previous \gls{sota} with over \textbf{$27\times$} computational decrease and over \textbf{$2\times$} better memory efficiency compared to the previous \gls{sota}. These results highlight FEMBA's versatility for high-performance applications and resource-restrictive scenarios.
\end{itemize}

By emphasizing both effectiveness and efficiency, FEMBA provides a step toward accessible, low-cost analytics in healthcare contexts. The remainder of this paper details the methodology and experiments, highlighting how state-space-based approaches can offer a compelling alternative to attention-driven architectures in \gls{eeg} analysis.

\section{Background}\label{sec:related}
This section provides an overview of the Temple University Hospital EEG (TUEG) Corpus and its labeled subsets, then reviews recent advances in EEG foundation models. We focus on the computational challenges faced by Transformer-based approaches, discuss the motivation for \glspl{ssm}, and examine Mamba-based solutions. Finally, we introduce how our proposed \emph{FEMBA} architecture builds upon these insights.

\subsection{Temple University Hospital EEG (TUEG)}\label{sec:TUEG}
The TUEG Corpus~\cite{obeid2016temple}, is one of the largest publicly available clinical EEG repositories. It contains over $26{,}000$ EEG recordings drawn from more than $14{,}000$ patients, spanning pediatric to geriatric populations and encompassing a variety of neurological conditions. In total, TUEG covers approximately $21{,}000$ hours of EEG data. Such diversity in demographics and pathologies provides a robust environment for learning general EEG representations.

\subsection{Key Labeled Subsets: TUAB, TUAR, and TUSL}\label{sec:datasets_finetune}
The TUEG dataset offers subsets of labeled datasets, such as the Temple University Hospital Abnormal EEG (TUAB), Artifact (TUAR) and Slowing (TUSL) Corpus~\cite{obeid2016temple}. TUAB offers annotated recordings labeled as \emph{normal} or \emph{abnormal}. TUAB has $2,329$ subjects and relatively balanced classes and TUAB serves as a strong benchmark for clinical diagnostics. The TUAR dataset contains annotations for various artifacts (e.g., eye blinks, muscle artifacts) in single-channel or multi-channel settings and has $213$ subjects. While in TUSL, the focus shifts to detecting and classifying \emph{slowing} events, seizures, complex background, and normal EEG. This 4-class classification task (slowing, seizure, complex, normal) consists of $1000$ subjects. Table~\ref{tab:dataset_summary} summarizes these three labeled subsets used in our experiments.

\begin{table}[t]
    \centering
    \caption{Summary of Datasets Used}
    \label{tab:dataset_summary}
    \begin{tabular}{lcc}
        \hline
        \textbf{Dataset} & \textbf{\# Subjects} & \textbf{Task} \\
        \hline
        TUEG & $14{,}987$ & Pre-training \\
        TUAB & $2{,}329$ & Abnormal vs. Normal \\
        TUAR & 213 & Artifact Detection \\
        TUSL & $38$ & Slowing Events \\
        \hline
    \end{tabular}
    \vspace{-0.5cm}
\end{table}

\subsection{Related works}
\paragraph*{Foundation Models in \gls{eeg}}
Foundation models have gained significant traction in NLP (e.g., DeepSeek~\cite{wu2024deepseek}) and computer vision (e.g., Molmo~\cite{deitke2024molmo}), motivating interest in their application to \gls{eeg}. However, these models are typically tailored for structured data such as text or images, raising challenges when dealing with the temporal complexity and biological variability of \gls{eeg} signals~\cite{cui2024toward}. Early \gls{eeg}-focused foundation models like BENDR~\cite{kostas_bendr_2021} employed contrastive learning yet faced scalability issues. Neuro-GPT~\cite{cui_neuro-gpt_2024} introduced autoregressive masking and reported gains in motor imagery classification, while LaBraM~\cite{jianglarge} and EEGFormer~\cite{chen2024eegformer} refined masked modeling methods across multiple datasets, achieving balanced accuracies above 80\% on abnormal \gls{eeg} detection. Despite these advancements, most prior approaches rely on Transformer architectures with $\mathcal{O}(N^2)$ complexity as a function of the sequence length, limiting their viability for continuous or large-scale \gls{eeg} monitoring.

\paragraph*{From Traditional Methods to State Space Models}
Conventional \gls{eeg} analysis often used machine learning algorithms such as \glspl{svm} and \gls{lda}, complemented by smaller deep networks like EEGNet~\cite{lawhern2018eegnet} and DeepConvNet~\cite{schirrmeister2017deep}. While these approaches offered interpretability and efficiency for relatively constrained tasks, they required extensive feature engineering and did not always generalize well to diverse patient populations. Transformer-based methods~\cite{jianglarge,chen2024eegformer,cui_neuro-gpt_2024,kostas_bendr_2021} later tackled the challenge of capturing long-range dependencies, though their substantial computational and memory demands may hinder real-world deployment.

\glspl{ssm} have gained interest for time-series analysis as they evolve a hidden state over time according to a simple linear dynamical system. In continuous form
\begin{equation*} 
\mathbf{h}'(t) = A\,\mathbf{h}(t) + B\,\mathbf{x}(t), \quad \mathbf{y}(t) = C\,\mathbf{h}(t), 
\end{equation*}
where $\mathbf{h}(t)$ is the hidden state, $\mathbf{x}(t)$ is the input, $\mathbf{y}(t)$ is the output, and $\{A, B, C\}$ are system matrices governing state evolution and output generation. Although these equations describe a continuous process, many implementations rely on discrete versions for efficient training in deep learning frameworks.

\paragraph*{Wearable and Edge Constraints}
Limited battery life, on-board memory, and compute resources characterize many real-world EEG applications, especially wearable devices~\cite{ingolfsson_brainfusenet_2024}. Applications like continuous epilepsy detection add real-time considerations and demand low false-alarm rates~\cite{ingolfsson_minimizing_2024}. The quadratic scaling of transformer-based methods often proves impractical under these constraints. In contrast, architectures based on state-space principles—owing to linear time and memory complexity—can better meet edge-computing requirements.

\paragraph*{Mamba-Based Approaches for \gls{eeg}}
A notable example of such \gls{ssm} is Mamba~\cite{gu2023mamba}, which applies a Zero-Order Hold (ZOH) scheme~\cite{pechlivanidou2022zero} to discretize the \gls{ssm}. Under a sampling interval $\Delta$, the continuous matrices $A,B$ map to discrete counterparts $A_d$ and $B_d$. Mamba further integrates a selective gating mechanism to modulate the hidden state update in a data-dependent manner. As a result, it achieves \emph{linear} complexity in sequence length, contrasting with the $\mathcal{O}(N^2)$ complexity of transformers. Bi-Mamba+~\cite{liang_bi-mamba_2024} extends Mamba by processing the input sequence in forward and backward directions, subsequently merging the two representations (e.g., via summation or gating). Recent work has begun to explore Mamba’s potential in \gls{eeg} analysis. \emph{Mentality}~\cite{Panchavati_mentality_2024} employed Mamba with a masked reconstruction scheme on TUSZ v2.0.1, improving seizure detection area under the ROC curve (AUROC) from 0.64 to 0.72. \emph{EEGMamba}~\cite{gui_eegmamba_2024} adopted a multi-task strategy by integrating Spatio-Temporal-Adaptive modules and Mixture-of-Experts heads, achieving above 98\% accuracy on the Siena dataset and around 97\% on CHB-MIT. Despite these early successes, challenges for Mamba-based models remain—especially regarding robust spatial-channel modeling for varying electrode montages and the need for domain-generalizable representations.

In this work, \textbf{our FEMBA} builds upon Mamba’s efficient state-space design by integrating large-scale self-supervised pre-training with bidirectional state updates, FEMBA aims to deliver strong accuracy on various \gls{eeg} downstream tasks while maintaining linear scaling (with regards to sequence length) suitable for resource-limited devices.

\section{Methodology}
\label{sec:methods}
In this section, we describe the proposed \emph{Foundational EEG Mamba + Bidirectional Architecture} (FEMBA) and its training procedures. Next, we outline our self-supervised pre-training scheme and finally, we explain our fine-tuning strategy, including two alternative classifier architectures and multiple downstream tasks (abnormal EEG detection, artifact recognition, and slowing event classification).

\subsection{Foundational EEG Mamba + Bidirectional Architecture (FEMBA)}
Our proposed FEMBA architecture is designed in four model sizes: Tiny, Base, Large, and Huge, with parameter sizes ranging from 7.8 million (Tiny) to 386 million (Huge), aligning with model sizes commonly explored in the literature~\cite{jianglarge,chen2024eegformer}. The primary distinction across these variants lies in the number of Bi-Mamba blocks and the embedding dimension, which is controlled by the 2D Convolution in the Tokenizer, as illustrated in Fig~\ref{fig:femba_architecture}. Specifically, the embedding dimensions for these configurations are as follows: the Tiny model uses two blocks and an embedding size of 35 ($(2,35)$); the Base model employs a configuration of $(12,35)$; the Large model adopts $(4,79)$; and the Huge model features $(20,79)$. Notably, the hidden state size across all configurations remains fixed at 80.

Each Bidirectional Mamba block processes the input sequence forward and backward, with the backward pass operating on a reversed sequence copy. Their outputs, potentially transformed as shown in Figure 1, are combined (e.g., summed) to integrate bidirectional context before further processing. A residual connection facilitates gradient flow.

During training, we utilize a layer-wise learning rate decay~\cite{ishii2017layer} with a fixed decay factor of $0.75$, progressively reducing the learning rate from the deeper blocks to the earlier ones.

\subsection{Self-Supervised Pre-training}
\label{subsec:pretraining}
We pre-train FEMBA on the TUEG dataset, as detailed in Section~\ref{sec:TUEG}. To prevent data leakage between pre-training and downstream tasks, we use a version of TUEG where subjects present in TUSL, TUAR, or TUAB have been filtered out. During pre-training, we adopt a self-supervised masked training strategy designed to enable FEMBA to learn robust, general-purpose representations of \gls{eeg} signals. This involves randomly masking a subset (60\%) of the input patches and training the model to reconstruct the missing patches, thereby compelling the encoder to capture meaningful spatiotemporal structures within the \gls{eeg} data.

\paragraph{Signal Normalization and Patch Embedding.}
We begin by representing each raw EEG recording as a tensor $x \in \mathbb{R}^{C \times T}$, where $C$ is the number of channels and $T$ is the temporal length (in samples). To reduce the influence of outliers, we apply quartile-based normalization~\cite{bedeeuzzaman2012automatic}, scaling each channel by its interquartile range (IQR):
\begin{equation*}
    x_{\text{norm}} = \frac{x - q_{\text{lower}}}{(q_{\text{upper}} - q_{\text{lower}}) + 1 \times 10^{-8}}.
\end{equation*}
We then segment $x_{\text{norm}}$ into bi-dimensional patches of size $p \times q$ (e.g., $4$ channels $\times$ $32$ samples). A 2D convolution projects these patches into an embedding space $\mathbf{X}_{\text{embed}} \in \mathbb{R}^{d \times C' \times T'}$, followed by learnable positional embeddings to maintain ordering across patch tokens.

\paragraph{Random Masking and Encoder.}
Next, we apply random masking to $60\%$ of the embedded patches, setting their representations to zero. This relatively high masking ratio ensures that the model must rely on contextual cues from unmasked segments to infer the missing patches. The masked embeddings, $\mathbf{X}_{\text{masked}}$, are then fed into the FEMBA encoder.

\paragraph{Decoder and Smooth L1 Reconstruction Loss.}
A lightweight decoder of two convolutional layers and a final linear projection attempts to reconstruct the original patches from the encoder outputs. We compute a Smooth L1 loss~\cite{girshick2015fastrcnn} only over the masked patches:
\begin{equation*}
    \text{SmoothL1}(\hat{x}, x) =
    \begin{cases}
        0.5 \,(x - \hat{x})^2, & \text{if } |x - \hat{x}| < \beta, \\
        |x - \hat{x}| - 0.5, & \text{otherwise},
    \end{cases}
\end{equation*}
\begin{equation*}
    \text{masked\_loss} = \frac{1}{|\mathcal{M}|} \sum_{i \in \mathcal{M}} \text{SmoothL1}(\hat{x}_i, x_i),
\end{equation*}
where $\mathcal{M}$ is the set of masked patch indices.

\subsection{Fine-Tuning}
\label{subsec:fine-tuning}

\subsubsection{Classifier Architectures.}
Following pre-training, the decoder is discarded and the Bi-Mamba encoder is repurposed as a feature extractor for downstream tasks. Two classification heads are explored:
\begin{enumerate}
    \item \textbf{Linear Classifier}: A small stack of fully connected layers (with GELU activations) outputs class probabilities. This design has a low parameter footprint ($\sim 0.5$\,M).
    \item \textbf{Mamba-Enhanced Classifier}: We add one more Mamba block before the final linear layer, enabling additional temporal modeling. This often improves accuracy in tasks with complex temporal dependencies but adds a slight increase in parameters (up to $0.7$\,M).
\end{enumerate}
\subsubsection{Downstream tasks}
We assess FEMBA on three downstream tasks using the datasets described in Section~\ref{sec:datasets_finetune}. For the TUAB dataset this consists of a binary classification (normal vs. abnormal), using the pre-defined train-test split. In TUSL, the task is a four-class classification task (slowing, seizure, complex, normal), Since the TUSL dataset lacks a predefined test split, we adopt an 80/10/10 randomized training/validation/test split. For TUAR we experiment with four versions of a downstream task based on the labeling scheme in in~\cite{ingolfsson2022energy}, they are described as the following:
\begin{itemize}
    \item \textbf{Binary Classification (BC)}: Label a window as \emph{artifact} if \emph{any} of the 13 artifact types is present on any channel; otherwise \emph{normal}.
    \item \textbf{Multilabel Classification (MC)}: Perform channel-wise artifact detection as a set of independent binary classifications, allowing multiple artifact types per window/channel.
    \item \textbf{Multiclass--Multioutput Classification (MMC)}: Discriminate between 13 artifact types for each channel, thus providing a more granular classification per channel.
    \item \textbf{Multiclass Classification (MCC)}: Restrict to 5 artifact types in a single-label setting, ignoring windows with combinations of artifacts (less than 5\% of data). This setting aligns closely with the protocol described by EEGFormer~\cite{chen2024eegformer}.
\end{itemize}
As the TUAR dataset also lacks a predefined test split, we similarly use an 80/10/10 randomized training/validation/test split.

\section{Results}\label{sec:results}
This section demonstrates that \textbf{FEMBA} consistently achieves \gls{soa} or near-\gls{soa} performance on diverse EEG benchmarks (TUAB, TUAR, and TUSL), while using significantly fewer FLOPs and less memory compared to recent \gls{soa} self-supervised Transformer-based methods. We provide quantitative accuracy metrics and efficiency analyses, which underscores FEMBA’s suitability for large-scale clinical or wearable EEG systems. For specific training details we fine-tune all layers (encoder + classifier) end-to-end using the Adam optimizer (initial learning rate of $1 \times 10^{-4}$) with cosine decay scheduling. Early stopping is employed based on validation loss to mitigate overfitting.

\begin{table}[htbp]
    \centering
    \caption{Performance Comparison on TUAB}
    \label{tab:results_tuab}
    \resizebox{\columnwidth}{!} {
    \setlength{\tabcolsep}{6pt} 
    \begin{tabular}{@{}lcccc@{}}
        \toprule
        \textbf{Model} & \textbf{Model Size} & \textbf{Bal. Acc. (\%)} & \textbf{AUPR} & \textbf{AUROC} \\ 
        \midrule
        \textbf{Supervised Models} \\
        SPaRCNet & 0.8M & 78.96 $\pm$ 0.18 & 0.8414 $\pm$ 0.0018 & 0.8676 $\pm$ 0.0012 \\
        ContraWR & 1.6M & 77.46 $\pm$ 0.41 & 0.8421 $\pm$ 0.0140 & 0.8456 $\pm$ 0.0074 \\
        CNN-Transformer & 3.2M & 77.77 $\pm$ 0.22 & 0.8433 $\pm$ 0.0039 & 0.8461 $\pm$ 0.0013 \\
        FFCL & 2.4M & 78.48 $\pm$ 0.38 & 0.8448 $\pm$ 0.0065 & 0.8569 $\pm$ 0.0051 \\
        ST-Transformer & 3.2M & 79.66 $\pm$ 0.23 & 0.8521 $\pm$ 0.0026 & 0.8707 $\pm$ 0.0019 \\
        \midrule
        \textbf{Self-superv. Models} \\
        BENDR & 0.39M & 76.96 $\pm$ 3.98 &  & 0.8397 $\pm$ 0.0344 \\
        BrainBERT & 43.2M & - & 0.8460 $\pm$ 0.0030 & 0.8530 $\pm$ 0.0020 \\
        EEGFormer-Small & 1.9M & - & 0.8620 $\pm$ 0.0050 & 0.8620 $\pm$ 0.0070 \\
        EEGFormer-Base & 2.3M & - & 0.8670 $\pm$ 0.0020 & 0.8670 $\pm$ 0.0030 \\
        EEGFormer-Large & 3.2M & - & 0.8720 $\pm$ 0.0010 & 0.8760 $\pm$ 0.0030 \\
        BIOT & 3.2M & 79.59 $\pm$ 0.57 & 0.8692 $\pm$ 0.0023 & 0.8815 $\pm$ 0.0043 \\
        EEG2Rep & - & 80.52 $\pm$ 2.22 &  & 0.8843 $\pm$ 0.0309 \\
        LaBraM-Base & 5.8M & 81.40 $\pm$ 0.19 & 0.8965 $\pm$ 0.0016 & 0.9022 $\pm$ 0.0009 \\
        LaBraM-Large & 46M & 82.26 $\pm$ 0.15 & 0.9130 $\pm$ 0.0005 & 0.9127 $\pm$ 0.0005 \\
        LaBraM-Huge & 369M & 82.58 $\pm$ 0.11 & 0.9204 $\pm$ 0.0011 & 0.9162 $\pm$ 0.0016 \\
        \midrule
        \textbf{FEMBA-Base} & 47.7M  &81.05 $\pm$ 0.14 &0.8894 $\pm$0.0050  & 0.8829 $\pm$ 0.0021  \\
        \textbf{FEMBA-Large} & 77.8M &  81.47 $\pm$ 0.11 & 0.8992 $\pm$ 0.0007  & 0.8856 $\pm$ 0.0004 \\
        \textbf{FEMBA-Huge} & 386M  & 81.82 $\pm$ 0.16 & 0.9005 $\pm$ 0.0017 & 0.8921 $\pm$ 0.0042  \\
        \bottomrule
   \end{tabular}}
   
\end{table}
\begin{table}[b]
\vspace{-0.5cm}
    \centering
    \caption{Model Comparison of FLOPs, Parameters, and Peak Memory Usage}
    \label{tab:model_comparison}
    \setlength{\tabcolsep}{6pt} 
    \renewcommand{\arraystretch}{1.2} 
    \small 
    \begin{tabular}{@{}lccc@{}}
    \toprule
    \textbf{Model} & \textbf{FLOPs} & \textbf{Parameters} & \textbf{Memory (MB)} \\ 
    \midrule
    EEGFormer-Small     & 21.06B & 1.9M     & 44.63 \\ 
    EEGFormer-Base      & 26.20B & 2.3M     & 71.32 \\ 
    EEGFormer-Large     & 36.46B & 3.2M     & 108.02 \\ 
    LaBraM-Base         & 4.42B  & 5.8M     & 757.38 \\ 
    LaBraM-Large        & 27.79B & 46M      & 1371.92 \\ 
    LaBraM-Huge         & 202.17B & 369M    & 2758.42 \\ 
    \textbf{FEMBA-Tiny} & 1.31B  & 7.8M     & 53.36 \\ 
    \textbf{FEMBA-Base} & 7.52B  & 47.7M    & 240.50 \\ 
    \textbf{FEMBA-Large} & 12.48B & 77.8M   & 548.71 \\ 
    \textbf{FEMBA-Huge} & 58.74B & 386M     & 1886.17 \\ 
    \bottomrule
    \end{tabular}
\end{table}
\begin{table*}[t]
    \centering
    \caption{Detailed Results on TUAR Across Four Classification Protocols}
    \label{tab:results_tuar}
    \small 
    \begin{tabular}{@{}lccccccc@{}}
    \toprule
    \textbf{Model} & \textbf{Model Size} & \multicolumn{2}{c}{\textbf{BC}} & \multicolumn{2}{c}{\textbf{MC}} & \multicolumn{2}{c}{\textbf{MMC}} \\
    \cmidrule(lr){3-4} \cmidrule(lr){5-6} \cmidrule(lr){7-8}
     &  & \textbf{AUROC} & \textbf{AUPR} & \textbf{AUROC} & \textbf{AUPR} & \textbf{AUROC} & \textbf{AUPR} \\
    \midrule
    \textbf{FEMBA-Tiny}  & 7.8M  
     & 0.937 $\pm$ 0.008 & 0.912 $\pm$ 0.010  
     & 0.887 $\pm$ 0.029 & 0.645 $\pm$ 0.024  
     & 0.893 $\pm$ 0.005 & 0.504 $\pm$ 0.013 
    \\
    \textbf{FEMBA-Base}  & 47.7M 
     & 0.949 $\pm$ 0.002 & 0.932 $\pm$ 0.001 
     & 0.909 $\pm$ 0.004 & 0.634 $\pm$ 0.016 
     & 0.888 $\pm$ 0.004 & 0.518 $\pm$ 0.002 
    \\
    \textbf{FEMBA-Large} & 77.8M  
     & 0.944 $\pm$ 0.003 & 0.913 $\pm$ 0.016 
     & 0.899 $\pm$ 0.006 & 0.608 $\pm$ 0.011 
     & 0.878 $\pm$ 0.020 & 0.516 $\pm$ 0.008 
    \\
    \bottomrule
    \end{tabular}
    \vspace{-0.2cm}
\end{table*}

\begin{table}[h!]
\centering
\caption{Performance Comparison across TUAR, TUSL}
\label{tab:results_tusl_neonate}
\resizebox{\columnwidth}{!} {
\setlength{\tabcolsep}{2pt} 
\begin{tabular}{@{}lccccccc@{}}
\toprule
\textbf{Model} & \textbf{Method Size} & \multicolumn{2}{c}{\textbf{TUAR}} & \multicolumn{2}{c}{\textbf{TUSL}} \\
\cmidrule(lr){3-4} \cmidrule(lr){5-6}
 &  & \textbf{AUROC} & \textbf{AUPR} & \textbf{AUROC} & \textbf{AUPR} \\
\midrule
EEGNet       & -  & 0.752 $\pm$ 0.006 & 0.433 $\pm$ 0.025 & 0.635 $\pm$ 0.015 & 0.351 $\pm$ 0.006  \\
TCN          & - & 0.687 $\pm$ 0.011 & 0.408 $\pm$ 0.009 & 0.545 $\pm$ 0.009 & 0.344 $\pm$ 0.001  \\
EEG-GNN      & - & 0.837 $\pm$ 0.022 & 0.488 $\pm$ 0.015 & 0.721 $\pm$ 0.009 & 0.381 $\pm$ 0.004  \\
GraphS4mer   & -  & 0.833 $\pm$ 0.006 & 0.461 $\pm$ 0.024 & 0.632 $\pm$ 0.017 & 0.359 $\pm$ 0.001  \\
BrainBERT    & 43.2M  & 0.753 $\pm$ 0.012 & 0.350 $\pm$ 0.014 & 0.588 $\pm$ 0.013 & 0.352 $\pm$ 0.003  \\
EEGFormer-Small  & 1.9M  & 0.847 $\pm$ 0.013 & 0.488 $\pm$ 0.012 & 0.683 $\pm$ 0.018 & 0.397 $\pm$ 0.011  \\
EEGFormer-Base  & 2.3M & 0.847 $\pm$ 0.014 & 0.483 $\pm$ 0.026 & 0.713 $\pm$ 0.010 & \textbf{0.393 $\pm$ 0.003}  \\
EEGFormer-Large  & 3.2M & 0.852 $\pm$ 0.004 & 0.483 $\pm$ 0.014 & 0.679 $\pm$ 0.013 & 0.389 $\pm$ 0.003  \\
\midrule
\textbf{FEMBA-Tiny}  & 7.8M  & \textbf{0.918 $\pm$ 0.003} & 0.518 $\pm$ 0.002 & 0.708 $\pm$ 0.005 & 0.277 $\pm$ 0.007 \\
\textbf{FEMBA-Base}  & 47.7M & 0.900 $\pm$ 0.010 & \textbf{0.559 $\pm$ 0.002}& \textbf{0.731 $\pm$ 0.012} & 0.289 $\pm$ 0.009\\
\textbf{FEMBA-Large} & 77.8M  & 0.915 $\pm$ 0.003 & 0.521 $\pm$ 0.001 &0.714 $\pm$ 0.007 & 0.282 $\pm$ 0.010 \\

\bottomrule
\end{tabular}
}
\vspace{-0.4cm}
\end{table}
\subsection{Pre-training results}
Our FEMBA model variants are initially pretrained to reconstruct both masked and unmasked sections of the signal, as detailed in Section~\ref{subsec:pretraining}. All variants demonstrated strong reconstruction capabilities for both masked and unmasked portions of the signal. This is illustrated in Fig~\ref{fig:reconstruct}, where the FEMBA-Base model successfully reconstructs a masked signal. The training and validation loss during pretraining were closely aligned for all variants, with example loss values for FEMBA base of $0.122$ (Train) and $0.217$ (Validation).
\input{Figures/reconstruct}

\subsection{TUAB: Abnormal EEG Detection}
Table~\ref{tab:results_tuab} summarizes TUAB results, where the task is to classify recordings as \emph{normal} or \emph{abnormal}. All \textbf{FEMBA} variants outperform the supervised models, with \textbf{FEMBA-Huge} attaining a balanced accuracy of 81.82\% , approaching LaBraM-Large/Huge~\cite{jianglarge} (82.26\%--82.58\%) but with around \(\mathbf{70\%}\) fewer FLOPs than LaBraM-Huge (see Table~\ref{tab:model_comparison}). Moreover, FEMBA outperforms EEGFormer-Large~\cite{chen2024eegformer} in AUROC (0.8921 vs. 0.8760). This underscores that our near-linear Mamba-based encoder can rival top Transformer architectures without incurring the quadratic attention cost.

\subsection{TUAR: Artifact Detection}
We next evaluate FEMBA on the Temple University Hospital Artifact (TUAR) dataset using four classification protocols of increasing label complexity: \textbf{BC} (binary), \textbf{MC} (multilabel), \textbf{MMC} (multiclass--multioutput), and \textbf{MCC} (multiclass single-label). Table~\ref{tab:results_tuar} details the performance of three FEMBA variants:

\paragraph{Binary Classification (BC).}
Even our smallest \textbf{FEMBA-Tiny} (7.8M parameters) achieves an AUROC of 0.937 and AUPR of 0.912, signaling robust artifact vs.\ normal discrimination. Scaling to \textbf{FEMBA-Base} boosts AUROC to 0.949 and AUPR to 0.932—about a 1.2\% gain in AUROC at a modest increase in parameters.

\paragraph{Multilabel (MC) \& Multiclass–Multioutput (MMC).}
Channel-wise artifact detection (MC) sees AUROCs of up to 0.909, while the more fine-grained MMC reaches 0.893. Notably, \textbf{FEMBA-Tiny} slightly outperforms the Base and Large variants in MMC (0.893 vs.\ 0.888/0.878), showcasing that a lean state-space model can excel even in complex multi-artifact labeling.

\paragraph{Multiclass Classification (MCC).}
Restricting windows to a single artifact type yields the highest AUROC (up to 0.918 for FEMBA-Tiny). Meanwhile, \textbf{FEMBA-Large} achieves 0.915 AUROC and the highest AUPR (0.521). As reported in Table~\ref{tab:results_tusl_neonate}, FEMBA also surpasses EEGFormer-l~\cite{chen2024eegformer} (0.852 AUROC) under a comparable MCC protocol, demonstrating a SoA result at a fraction of the Transformer’s computational cost.

\subsection{TUSL (Slowing Event Classification).}
Table~\ref{tab:results_tusl_neonate} indicates that \textbf{FEMBA-Base} achieves 0.731~AUROC, surpassing EEGFormer-Small/Large by 4.8\%–5.2\% absolute (0.683/0.679), and slightly outperforming EEGFormer-Base (0.713). However, FEMBA’s AUPR (0.289) trails the best EEGFormer-Large AUPR (0.389) by about 10 percentage points, likely due to class imbalance. Despite this, FEMBA demonstrates these results at a significantly lower computational cost, as detailed in Section~\ref{subsec:efficiency}.

\subsection{Efficiency Analysis: FLOPs, Parameters, and Memory}\label{subsec:efficiency}
Practical considerations—such as floating-point operations (FLOPs), parameter counts, and peak memory usage—are critical in determining the feasibility of real-world or continuous EEG monitoring. Table~\ref{tab:model_comparison} provides a comparison of major Transformer baselines (EEGFormer, LaBraM) and our FEMBA models across these metrics.

\begin{figure*}[t]
\centering
\begin{minipage}{.49\textwidth}
\begin{tikzpicture}

\definecolor{darkgreen}{RGB}{0,100,0}
\definecolor{gray}{RGB}{128,128,128}
\definecolor{indianred1967882}{RGB}{196,78,82}
\definecolor{lightgray204}{RGB}{204,204,204}
\definecolor{mediumseagreen85168104}{RGB}{85,168,104}
\definecolor{steelblue76114176}{RGB}{76,114,176}

\begin{axis}[
height=6cm,
width=\columnwidth,
legend cell align={left},
legend style={
  fill opacity=0.8,
  draw opacity=1,
  text opacity=1,
  at={(0.03,0.97)},
  anchor=north west,
  draw=lightgray204
},
tick align=outside,
tick pos=left,
x grid style={black!20},
xmajorgrids,
xlabel near ticks,
xmin=-0.43, xmax=2.43,
xtick style={color=black},
xtick={0,1,2},
xticklabels={Tiny,Base,Huge},
y grid style={black!20},
ylabel={FLOPs (Billion)},
ymajorgrids,
ymin=0, ymax=222.31,
ytick style={color=black}
]
\draw[draw=black,fill=steelblue76114176,very thin] (axis cs:-0.3,0) rectangle (axis cs:-0.1,0);
\addlegendimage{ybar,ybar legend,draw=black,fill=steelblue76114176,very thin}
\addlegendentry{LaBraM}

\draw[draw=black,fill=steelblue76114176,very thin] (axis cs:0.7,0) rectangle (axis cs:0.9,27.8);
\draw[draw=black,fill=steelblue76114176,very thin] (axis cs:1.7,0) rectangle (axis cs:1.9,202.2);
\draw[draw=black,fill=mediumseagreen85168104,very thin] (axis cs:-0.1,0) rectangle (axis cs:0.1,1.3);
\addlegendimage{ybar,ybar legend,draw=black,fill=mediumseagreen85168104,very thin}
\addlegendentry{FEMBA}

\draw[draw=black,fill=mediumseagreen85168104,very thin] (axis cs:0.9,0) rectangle (axis cs:1.1,7.5);
\draw[draw=black,fill=mediumseagreen85168104,very thin] (axis cs:1.9,0) rectangle (axis cs:2.1,58.7);
\draw[draw=black,fill=indianred1967882,very thin] (axis cs:0.1,0) rectangle (axis cs:0.3,36.5);
\addlegendimage{ybar,ybar legend,draw=black,fill=indianred1967882,very thin}
\addlegendentry{EEGFormer}

\draw[draw=black,fill=indianred1967882,very thin] (axis cs:1.1,0) rectangle (axis cs:1.3,0);
\draw[draw=black,fill=indianred1967882,very thin] (axis cs:2.1,0) rectangle (axis cs:2.3,0);
\draw (axis cs:-0.2,0) ++(0pt,3pt) node[
  scale=0.7,
  anchor=south,
  text=black,
  rotate=0.0
]{N/A};
\draw (axis cs:0.8,25) ++(0pt,3pt) node[
  scale=0.7,
  anchor=south,
  text=black,
  rotate=0.0
]{27.8};
\draw (axis cs:1.8,202.2) ++(0pt,3pt) node[
  scale=0.7,
  anchor=south,
  text=black,
  rotate=0.0
]{202.2};
\draw (axis cs:0,1.3) ++(0pt,3pt) node[
  scale=0.7,
  anchor=south,
  text=black,
  rotate=0.0
]{\bfseries 1.3};
\draw (axis cs:1,7.5) ++(0pt,3pt) node[
  scale=0.7,
  anchor=south,
  text=black,
  rotate=0.0
]{\bfseries 7.5};
\draw (axis cs:2,58.7) ++(0pt,3pt) node[
  scale=0.7,
  anchor=south,
  text=black,
  rotate=0.0
]{\bfseries 58.7};
\draw (axis cs:0.2,36.5) ++(0pt,3pt) node[
  scale=0.7,
  anchor=south,
  text=black,
  rotate=0.0
]{36.5};
\draw (axis cs:1.2,0) ++(0pt,3pt) node[
  scale=0.7,
  anchor=south,
  text=black,
  rotate=0.0
]{N/A};
\draw (axis cs:2.2,0) ++(0pt,3pt) node[
  scale=0.7,
  anchor=south,
  text=black,
  rotate=0.0
]{N/A};
\draw[<-, draw=darkgreen, line width=1pt] (axis cs:0,31.3) -- (axis cs:0.15,60);

\draw (axis cs:0,59.3) node[
  scale=0.7,
  text=darkgreen,
  rotate=0.0
]{\bfseries 27.8x};
\draw[<-,draw=darkgreen, line width=1pt] (axis cs:1,37.5) -- (axis cs:0.8,47.8);
\draw (axis cs:0.9,59.3) node[
  scale=0.7,
  text=darkgreen,
  rotate=0.0
]{\bfseries 3.7x};
\draw[-,draw=darkgreen,line width=1pt] (axis cs:2.01,207.2) -- (axis cs:1.9,207.2);
\draw[<-,draw=darkgreen,line width=1pt] (axis cs:2,88.7) -- (axis cs:2,207.7);
\draw (axis cs:2.1,116.7) node[
  scale=0.7,
  text=darkgreen,
  rotate=0.0
]{\bfseries 3.4x};
\end{axis}

\end{tikzpicture}
\end{minipage}
\begin{minipage}{.5\textwidth}
\begin{tikzpicture}

\definecolor{darkgreen}{RGB}{0,100,0}
\definecolor{gray}{RGB}{128,128,128}
\definecolor{indianred1967882}{RGB}{196,78,82}
\definecolor{lightgray204}{RGB}{204,204,204}
\definecolor{mediumseagreen85168104}{RGB}{85,168,104}
\definecolor{steelblue76114176}{RGB}{76,114,176}

\begin{axis}[
height=6cm,
width=\columnwidth,
legend cell align={left},
legend style={
  fill opacity=0.8,
  draw opacity=1,
  text opacity=1,
  at={(0.03,0.97)},
  anchor=north west,
  draw=lightgray204
},
tick align=outside,
tick pos=left,
x grid style={black!20},
xmajorgrids,
xlabel near ticks,
xmin=-0.43, xmax=2.43,
xtick style={color=black},
xtick={0,1,2},
xticklabels={Tiny,Base,Huge},
y grid style={black!20},
ylabel={Peak GPU Memory Usage (MB)},
ymajorgrids,
ymin=0, ymax=3000,
ytick style={color=black}
]
\draw[draw=black,fill=steelblue76114176,very thin] (axis cs:-0.3,0) rectangle (axis cs:-0.1,0);
\addlegendimage{ybar,ybar legend,draw=black,fill=steelblue76114176,very thin}
\addlegendentry{LaBraM}

\draw[draw=black,fill=steelblue76114176,very thin] (axis cs:0.7,0) rectangle (axis cs:0.9,1371.9);
\draw[draw=black,fill=steelblue76114176,very thin] (axis cs:1.7,0) rectangle (axis cs:1.9,2758.4);
\draw[draw=black,fill=mediumseagreen85168104,very thin] (axis cs:-0.1,0) rectangle (axis cs:0.1,53.4);
\addlegendimage{ybar,ybar legend,draw=black,fill=mediumseagreen85168104,very thin}
\addlegendentry{FEMBA}

\draw[draw=black,fill=mediumseagreen85168104,very thin] (axis cs:0.9,0) rectangle (axis cs:1.1,240.5);
\draw[draw=black,fill=mediumseagreen85168104,very thin] (axis cs:1.9,0) rectangle (axis cs:2.1,1886.2);
\draw[draw=black,fill=indianred1967882,very thin] (axis cs:0.1,0) rectangle (axis cs:0.3,108);
\addlegendimage{ybar,ybar legend,draw=black,fill=indianred1967882,very thin}
\addlegendentry{EEGFormer}

\draw[draw=black,fill=indianred1967882,very thin] (axis cs:1.1,0) rectangle (axis cs:1.3,0);
\draw[draw=black,fill=indianred1967882,very thin] (axis cs:2.1,0) rectangle (axis cs:2.3,0);
\draw (axis cs:-0.2,0) ++(0pt,3pt) node[
  scale=0.7,
  anchor=south,
  text=black,
  rotate=0.0
]{N/A};
\draw (axis cs:0.8,1371.9) ++(0pt,3pt) node[
  scale=0.7,
  anchor=south,
  text=black,
  rotate=0.0
]{1.3k};
\draw (axis cs:1.8,2658.4) ++(0pt,3pt) node[
  scale=0.7,
  anchor=south,
  text=black,
  rotate=0.0
]{2.7k};
\draw (axis cs:0,53.4) ++(0pt,3pt) node[
  scale=0.7,
  anchor=south,
  text=black,
  rotate=0.0
]{53.4};
\draw (axis cs:1.01,240.5) ++(0pt,3pt) node[
  scale=0.7,
  anchor=south,
  text=black,
  rotate=0.0
]{\bfseries 240.5};
\draw (axis cs:2,1886.2) ++(0pt,3pt) node[
  scale=0.7,
  anchor=south,
  text=black,
  rotate=0.0
]{\bfseries 1.8k};
\draw (axis cs:0.2,108) ++(0pt,3pt) node[
  scale=0.7,
  anchor=south,
  text=black,
  rotate=0.0
]{\bfseries 108.0};
\draw (axis cs:1.2,0) ++(0pt,3pt) node[
  scale=0.7,
  anchor=south,
  text=black,
  rotate=0.0
]{N/A};
\draw (axis cs:2.2,0) ++(0pt,3pt) node[
  scale=0.7,
  anchor=south,
  text=black,
  rotate=0.0
]{N/A};
\draw[<-,draw=darkgreen, line width=1pt] (axis cs:0,350.4) -- (axis cs:0.2,458);
\draw (axis cs:0.1,550) node[
  scale=0.7,
  text=darkgreen,
  rotate=0.0
]{\bfseries 2x};
\draw[-,draw=darkgreen, line width=1pt] (axis cs:1.01,1471.9) -- (axis cs:0.9,1471.9);
\draw[<-,draw=darkgreen,line width=1pt] (axis cs:1,540.5) -- (axis cs:1,1481.9);
\draw (axis cs:1.1,840.5) node[
  scale=0.7,
  text=darkgreen,
  rotate=0.0
]{\bfseries 3.1x};
\draw[-,draw=darkgreen, line width=1pt] (axis cs:2.01,2858.4) -- (axis cs:1.9,2858.4);
\draw[<-,draw=darkgreen, line width=1pt] (axis cs:2,2186.2) -- (axis cs:2,2868.4);
\draw (axis cs:2.1,2386.2) node[
  scale=0.7,
  text=darkgreen,
  rotate=0.0
]{\bfseries 1.5x};
\end{axis}

\end{tikzpicture}

\end{minipage}
\caption{Comparison of LaBraM~\cite{jianglarge}, \textbf{FEMBA}, and EEGFormer~\cite{chen2024eegformer} in terms of computational inference (left) and memory usage (in megabytes, MB) (right)}\label{fig:comparison_inference_gpu}
\vspace{-0.5cm}
\end{figure*}
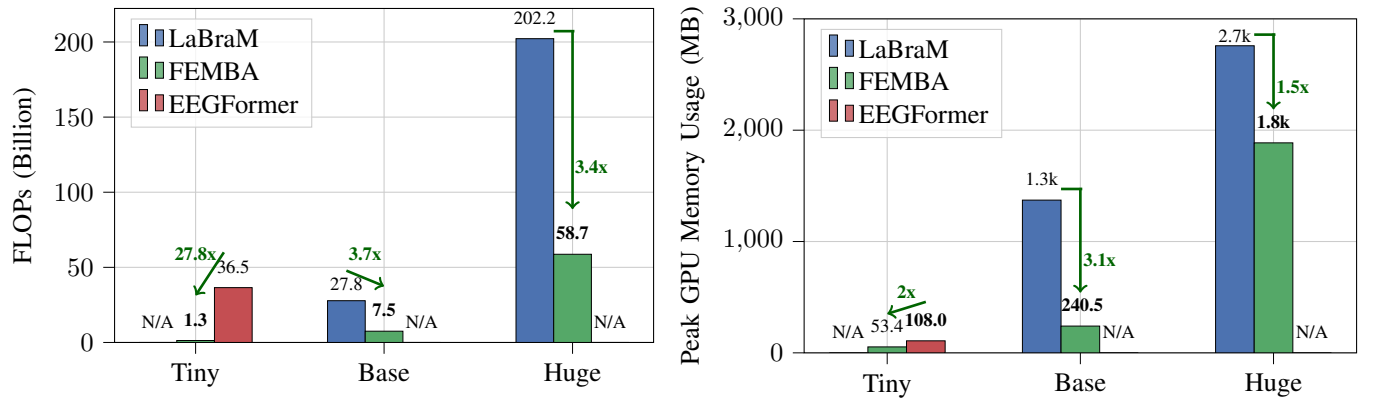

For \textbf{LaBraM}, FLOPs and memory usage are calculated using its publicly available code repository. For \textbf{EEGFormer}, these metrics are approximated based on the limited details available in the literature, as no official code has been released. To measure peak memory usage, we process a batch size of 8 through each model and record the maximum memory consumption. Despite these approximations, a clear trend is evident:

\textbf{FEMBA-Huge} (386M parameters) requires 58.74B FLOPs, nearly \(\mathbf{3.5\times}\) fewer FLOPs than LaBraM-Huge (202.17B) and 30\% less memory usage, yet achieves comparable TUAB accuracy (81.82\% vs.\ 82.58\%).  \textbf{FEMBA-Tiny} (7.8M) uses only 1.31B FLOPs—up to \(\mathbf{27\times}\) fewer than EEGFormer-Large—while still delivering SoA AUROC (e.g., 0.918 on TUAR MCC). Similarly \textbf{FEMBA-Base} runs at 7.52B FLOPs, roughly \(\mathbf{4\times}\) lower than EEGFormer-Large (36.46B FLOPs). A detailed visual comparison of these models is provided in Figure~\ref{fig:comparison_inference_gpu}.

\subsection{Discussion}
Overall, FEMBA consistently achieves SoA or near-SoA accuracy with substantially reduced computational cost. On TUAB, \emph{FEMBA-Huge} falls within 0.8--1.0\% absolute of LaBraM-Large/Huge in balanced accuracy but uses roughly \(\mathbf{70\%}\) fewer FLOPs than LaBraM-Huge. On TUAR, \emph{FEMBA-Tiny} (7.8M) outperforms EEGFormer-l by 6.6\% in AUROC under comparable MCC protocols. For TUSL, FEMBA-Base surpasses all EEGFormer variants by up to 4.8\% in AUROC.

These findings validate that a state-space modeling approach can match or exceed Transformer baselines without the prohibitive \(\mathcal{O}(N^2)\) scaling. Beyond computational efficiency, model interpretability is crucial for clinical adoption. While Transformer interpretability often relies on attention maps, FEMBA's state-space nature offers different potential mechanisms. Future work could analyze learned state dynamics or use sensitivity analysis to understand how temporal EEG patterns are processed, potentially offering more temporally grounded explanations for clinical events than attention. Exploring these SSM-specific techniques versus Transformer approaches, especially for providing clinically meaningful explanations, is an important avenue for future research to build trust and facilitate deployment.

Future work could also explore enhancements to boost FEMBA’s accuracy, such as refining its architecture or incorporating advanced regularization techniques. Furthermore, the robust representations from large-scale pre-training suggest FEMBA's potential for generalization to other domains like sleep staging or BCI, although evaluating this and the need for fine-tuning remains a valuable next step. Additionally, neonatal-focused pre-training could address domain shifts, while multi-modal integration may extend FEMBA’s applicability to a wider range of clinical scenarios. We conclude that FEMBA’s efficient design and robust performance establish it as a compelling alternative to Transformer-based EEG models for both large-scale and on-device applications.

\section{Conclusion}\label{ch:conclusion}
We introduced \textbf{FEMBA}, a novel self-supervised EEG framework grounded in bidirectional state-space modeling and pre-trained on over 21,000 hours of unlabelled clinical EEG. Our experiments across multiple downstream tasks (abnormal EEG detection, artifact recognition, slowing event classification, and neonatal seizure detection) demonstrate that FEMBA achieves near-Transformer performance while maintaining significantly lower computational complexity and memory requirements.

Notably, a \emph{tiny} 7.8M-parameter variant (FEMBA-Tiny) retains competitive accuracy on tasks such as artifact detection, showcasing the potential for real-time edge deployments. Nonetheless, certain domain shifts—such as neonatal vs.\ adult EEG—underscore the need for additional domain adaptation. Future work will explore these techniques and integrate multi-modal physiological signals for more robust clinical event detection. We believe FEMBA marks a key step toward delivering efficient, universal EEG foundation models that operate seamlessly from large hospital databases to low-power wearable devices.
\section*{Acknowledgment}
\vspace{-0.1cm}
This project is supported by the Swiss National Science Foundation under the grant number 193813 (PEDESITE project) and by the ETH Future Computing Laboratory (EFCL), financed by a donation from Huawei Technologies. We acknowledge ISCRA for awarding this project access to the LEONARDO supercomputer, owned by the EuroHPC Joint Undertaking, hosted by CINECA (Italy). This work was supported by a grant from the Swiss National Supercomputing Centre (CSCS) under project ID lp12 on Alps.
\bibliographystyle{IEEEtran}
\bibliography{bib}

\end{document}